\documentclass[11pt,letterpaper]{article}
\usepackage{ijcnlp2017}
\usepackage{times}
\usepackage{latexsym,amsmath,amssymb,amsfonts}
\usepackage{graphicx}
\usepackage{multirow,rotating}
\usepackage{paralist}
\usepackage{booktabs}

\ijcnlpfinalcopy

\title{End-to-End Task-Completion Neural Dialogue Systems}

\author{Xiujun Li$^{\dagger}$\quad Yun-Nung Chen$^{\star}$\quad Lihong Li$^{\dagger}$\quad Jianfeng Gao$^{\dagger}$\quad Asli Celikyilmaz$^\dagger$\\
  $^{\dagger}$Microsoft Research, Redmond, WA, USA\\
  $^{\star}$National Taiwan University, Taipei, Taiwan\\
  $^{\star}${\tt y.v.chen@ieee.org} \\
  $^{\dagger}${\tt \{xiul,lihongli,jfgao,aslicel\}@microsoft.com}
}

\date{}

\begin{document}

\maketitle

\begin{abstract}
One of the major drawbacks of modularized task-completion dialogue systems is that each module is trained individually, which presents several challenges. For example, downstream modules are affected by earlier modules, and the performance of the entire system is not robust to the accumulated errors. This paper presents a novel end-to-end learning framework for task-completion dialogue systems to tackle such issues.
Our neural dialogue system can directly interact with a structured database to assist users in accessing information and accomplishing certain tasks. The reinforcement learning based dialogue manager offers robust capabilities to handle noises caused by other components of the dialogue system.
Our experiments in a movie-ticket booking domain show that our end-to-end system not only outperforms modularized dialogue system baselines for both objective and subjective evaluation, but also is robust to noises as demonstrated by several systematic experiments with different error granularity and rates specific to the language understanding module\footnote{The source code is available at: \url{https://github/com/MiuLab/TC-Bot}.}.
\end{abstract}

\section{Introduction}
In the past decade, goal-oriented dialogue systems have been the most prominent component in today's virtual personal assistants, which allow users to speak naturally in order to accomplish tasks more efficiently. Traditional systems have a rather complex and modularized pipeline, consisting of a language understanding (LU) module, a dialogue manager (DM), and a natural language generation (NLG) component~\cite{rudnicky1999creating,zue2000juplter,zue2000conversational}.

Recent advances of deep learning have inspired many applications of neural models to dialogue systems. \newcite{wen2017network} and \newcite{bordes2017learning} introduced a network-based end-to-end trainable task-oriented dialogue system, which treated dialogue system learning as the problem of learning a mapping from dialogue histories to system responses, and applied an encoder-decoder model to train the whole system. However, the system is trained in a supervised fashion: not only does it require a lot of training data, but it may also fail to find a good policy robustly due to lack of exploration of dialogue control in the training data. 
\newcite{zhao2016towards} first presented an end-to-end reinforcement learning (RL) approach to dialogue state tracking and policy learning in the DM. This approach is shown to be promising when applied to the task-oriented dialogue problem of guessing the famous person a user thinks of. In the conversation, the agent asks the user a series of \textit{Yes/No} questions to find the correct answer. However, this simplified task may not generalize to practical problems due to the following: 

\begin{figure*}[t]
\centering
\includegraphics[width=\linewidth]{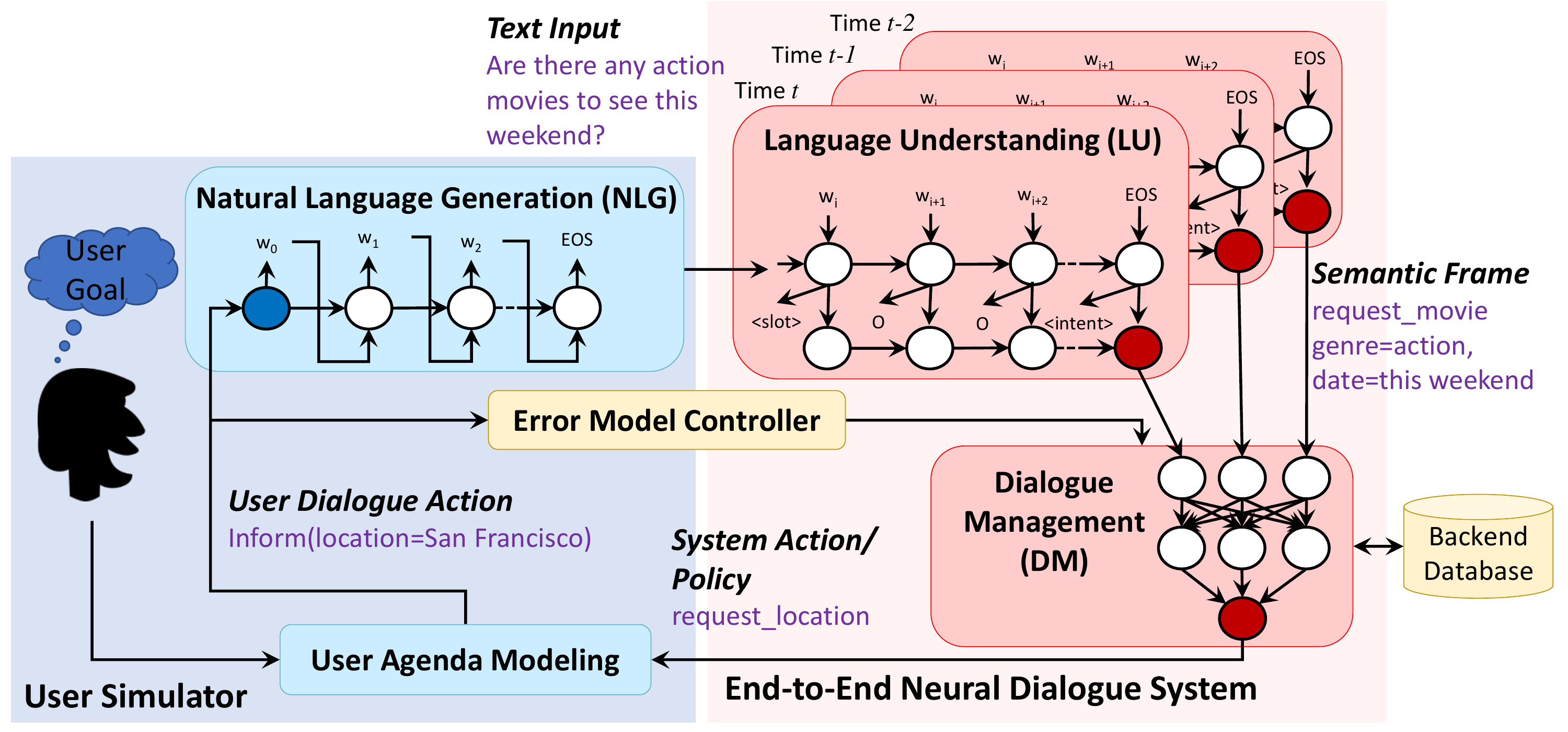}
\caption{Illustration of the end-to-end neural dialogue system: given user utterances, reinforcement learning is used to train all components in an end-to-end fashion.}
\label{fig:end2end}
\end{figure*}

\begin{compactenum}
\item \textbf{Inflexible question types} --- asking request questions is more natural and efficient than \textit{Yes/No} questions. For example, it is more natural and efficient for the system to ask ``\textit{Where are you located?}'' instead of ``\textit{Are you located in Palo Alto?}'', when there are a large number of possible values for the location slot.
\item \textbf{Poor robustness} --- the user answers are too simple to be misunderstood, so the system lacks the robustness against noise in real user utterances.
\item \textbf{User requests during dialogues} --- in a task-oriented dialogue, user may ask questions for selecting the preferred slot values. In a flight-booking example, user might ask ``\textit{What flight is available tomorrow?}''.
\end{compactenum}
For the second issue, \newcite{su2016continuously} briefly investigated the effect of dialogue action level semantic error rates on the dialogue performance. \newcite{lemon2007dialogue} compared policy transfer properties under different environments, showing that policies trained in high-noise conditions have better transfer properties than those trained in low-noise conditions. Recently, \newcite{dhingra2017end} proposed an end-to-end differentiable KB-Infobot to provide the solutions to the first two issues, but the last one remained unsolved. 


This paper addresses all three issues above by redefining the targeted system as a task-completion neural dialogue system. Our framework is more practical in that the information can be easily accessed by the user during the conversations, while the final goal of the system is to complete a task, such as movie-ticket booking. This paper is the first attempt of training a real-world task-completion dialogue system in an end-to-end fashion by leveraging supervised learning and reinforcement learning techniques. To further understand the robustness of reinforcement learning based dialogue systems, we conduct extensive experiments and quantitative analysis on a fine-grained level of LU errors, and provide meaningful insights on how the language understanding component impacts the overall performance of the dialogue system.

Our contributions are three-fold:
\begin{compactitem}
\item \textbf{Robustness} --- We propose a neural dialogue system with greater robustness by automatically selecting actions based on uncertainty and confusion by reinforcement learning. We also provide the first systematic analysis to investigate the impact of different types of natural language understanding errors on dialogue system performance. We show that slot-level errors have a greater impact on the system performance than intent-level ones, and that slot value replacement degrades the performance most. Our findings shed some light on how to design multi-task natural language understanding models (intent classification and slot labeling) in the dialogue systems with consideration of error control.
\item \textbf{Flexibility} --- The system is the first neural dialogue system that allows user-initiated behaviors during conversations, where the users can interact with the system with higher flexibility that is important in realistic scenarios.
\item \textbf{Reproducibility} --- We demonstrate how to evaluate RL dialogue agents using crowdsourced task-specific datasets and simulated users in an end-to-end fashion, guaranteeing  reproducibility and consistent comparisons of competing methods in an identical setting.
\end{compactitem}

\section{Proposed Framework}
The proposed framework\footnote{The source code is available at: \url{https://github.com/MiuLab/TC-Bot}} is illustrated in Figure~\ref{fig:end2end}. It includes a user simulator (left part) and a neural dialogue system (right part). In the user simulator, an agenda-based user modeling component based at the dialogue act level is applied to control the conversation exchange conditioned on the generated user goal, to ensure the user behaves in a consistent, goal-oriented manner. An NLG module is used to generate natural language texts corresponding to the user dialogue actions. In a neural dialogue system, an input sentence (recognized utterance or text input) passes through an LU module and becomes a corresponding semantic frame, and an DM, which includes a state tracker and policy learner, is to accumulate the semantics from each utterance, robustly track the dialogue states during the conversation, and generate the next system action.

\subsection{Neural Dialogue System}
\label{sec:sds}

\begin{figure}[t]
\begin{center}
\begin{small}
\begin{tabular}{|lccccc|}
\hline
{\bf W} &find&action&movies&this&weekend\\
& $\downarrow$&$\downarrow$&$\downarrow$&$\downarrow$&$\downarrow$\\
{\bf S} & O&B-genre&O&B-date&I-date\\
{\bf I} & \multicolumn{5}{l|} {find\_movie}\\
\hline
\end{tabular}
\end{small}
\caption{An example utterance with annotations of semantic slots in IOB format (S) and intent (I), B-date and I-date denote the date slot.}
\label{fig:iob}
\end{center}
\end{figure}

\paragraph{Language Understanding (LU):}
A major task of LU is to automatically classify the domain of a user query along with domain specific intents and fill in a set of slots to form a semantic frame. The popular IOB (in-out-begin) format is used for representing the slot tags, as shown in Figure~\ref{fig:iob}.
\begin{eqnarray*}
\vec{x} &=& w_1,..., w_n, \text{$<$EOS$>$} \\
\vec{y} &=& s_1,..., s_n, i_m
\end{eqnarray*}
where $\vec{x}$ is the input word sequence and $\vec{y}$ contains the associated slots, $s_k$, and the sentence-level intent $i_m$.
The LU component is implemented with a single LSTM, which performs intent prediction and slot filling simultaneously~\cite{hakkani2016multi,chen2016syntax}:
\begin{equation}
\vec{y} = \text{LSTM}(\vec{x})\,.
\end{equation}
The LU objective is to maximize the conditional probability of the slots and the intent $\vec{y}$ given the word sequence $\vec{x}$:
\begin{equation*}
p(\vec{y}\mid \vec{x}) = \left(\prod_i^n p(s_i\mid w_1, \ldots, w_i)\right)p(i_m\mid \vec{y}).
\end{equation*}
The weights of the LSTM model are trained using backpropagation to maximize the conditional likelihood of the training set labels. The predicted tag set is a concatenated set of IOB-format slot tags and intent tags; therefore, this model can be trained using all available dialogue actions and utterance pairs in our labeled dataset in a supervised manner.

\paragraph{Dialogue Management (DM):}
The symbolic LU output is passed to the DM in the dialogue act form (or semantic frame). The classic DM includes two stages, \emph{dialogue state tracking} and \emph{policy learning}. 
\begin{compactitem}
\item \textbf{Dialogue state tracking}: Given the LU symbolic output, such as \textsf{request(moviename; genre=action; date=this weekend)}, three major functions are performed by the state tracker: a symbolic query is formed to interact with the database to retrieve the available results; the state tracker will be updated based on the available results from the database and the latest user dialogue action; and the state tracker will prepare the state representation $s_t$ for policy learning.
\item \textbf{Policy learning}: The state representation for the policy learning includes the latest user action (e.g., \textsf{request(moviename; genre=action; date=this weekend)}), the latest agent action (\textsf{request(location)}), the available database results, turn information, and history dialogue turns, etc. Conditioned on the state representation $s_t$ from the state tracker, the policy $\pi$ is to generate the next available system action $a_t$ according to $\pi(s_t)$.
Either supervised learning or reinforcement learning can be used to optimize $\pi$.  Details about RL-based policy learning can be found in section~\ref{sec:e2e_rl}.
\end{compactitem}

Prior work used different implementation approaches summarized below. Dialogue state tracking is the process of constantly updating the state of the dialogue, and \newcite{lee2014extrinsic} showed that there is a positive correlation between state tracking performance and dialogue performance. Most production systems use manually designed heuristics, often based on rules, to update the dialogue states based on the highly confident output from LU. \newcite{williams2013dialog} formalized the tracking problem as a supervised sequence labeling task, where the input is LU outputs and the output is the true slot values, and the state tracker's results can be translated into a dialogue policy. \newcite{zhao2016towards} proposed to jointly train the state tracker and the policy learner in order to optimize the system actions more robustly. Instead of explicitly incorporating the state tracking labels, this paper learns the system actions with implicit dialogue states, so that the proposed DM can be more flexible and robust to the noise propagated from the previous components~\cite{su2016continuously,liu2017an}. A rule-based agent is employed to warm-start the system, via supervised learning on labels generated by the rules.  The system is then further trained end-to-end with RL, as explained in section~\ref{sec:e2e_rl}.

\subsection{User Simulation}
\label{sec:simulator}
In order to perform end-to-end training for the proposed neural dialogue systems, a user simulator is required to automatically and naturally interact with the dialogue system.
In the task-completion dialogue setting, the user simulator first generates a user goal. The agent does not know the user goal, but tries to help the user accomplish it in the course of conversations. Hence, the entire conversation exchange is around this goal implicitly. A user goal generally consists of two parts: 
\emph{inform\_slots} for slot-value pairs that serve as constraints from the user, and
\emph{request\_slots} for slots whose value the user has no information about, but wants to get the values from the agent during the conversation. The user goals are generated using a set of labeled conversational data.

\paragraph{User Agenda Modeling:}
\label{sec:user_agenda}
During the course of a dialogue, the user simulator maintains a compact, stack-like representation called \emph{user agenda}~\cite{schatzmann2009hidden}, where the user state $s_u$ is factored into an agenda $A$ and a goal $G$. The goal consists of constraints $C$ and request $R$. At each time-step $t$, the user simulator generates the next user action $a_{u,t}$ based on the current state $s_{u,t}$ and the last agent action $a_{m,t-1}$, and then updates the current status $s'_{u,t}$.

\paragraph{Natural Language Generation (NLG):}
Given the user's dialogue actions, the NLG module generates natural language texts. To control the quality of user simulation given limited labeled data, a hybrid approach including a template-based NLG and a model-based NLG is employed, where the model-based NLG is trained on the labeled dataset with a sequence-to-sequence model. It takes dialogue acts as input, and generates sentence sketch with slot placeholders via an LSTM decoder. Then a post-processing scan is performed to replace the slot placeholders with their actual values~\cite{wen2015semantically}. In the LSTM decoder, we apply beam search, which iteratively considers the top $k$ best sub-sentences when generating the next token.

In the hybrid model, if the user dialogue actions can be found in the predefined sentence templates, the template-based NLG is applied; otherwise, the utterance is generated by the model-based NLG. This hybrid approach allows a dialogue system developer to easily improve NLG by providing templates for sentences that the machine-learned model does not handle well.

\subsection{Error Model Controller}
\label{ssec:user_error}
When training or testing a policy based on semantic frames of user actions, an error model~\cite{schatzmann2007error} is introduced to simulate noises from the LU component, and noisy communication between the user and the agent in order to test the model robustness. Here, we introduce different levels of noises in the error model: one type of errors is at the \emph{intent} level, another is at the \emph{slot} level. For each level, there are more fine-grained noises. 

\paragraph{Intent-Level Error:}
At the intent level, we categorize all intents into three groups: 
\begin{compactitem}
\item \emph{Group 1}: general \emph{greeting}, \emph{thanks}, \emph{closing}, etc.
\item \emph{Group 2}: users may \emph{inform}, to tell the slot values (or constraints) to the agent, for example, \textsf{inform(moviename=`Titanic', starttime=`7pm')}.
\item \emph{Group 3}: users may \emph{request} information for specific slots. In a movie-booking scenario, users might ask ``\textsf{request(starttime; moviename=`Titanic')}''.
\end{compactitem}
In the specific task of movie-booking, for instance, there exist multiple \emph{inform} and \emph{request} intents, such as \textsf{request\_starttime}, \textsf{request\_moviename}, \textsf{inform\_starttime} and \textsf{inform\_moviename}, etc.
Based on the above intent categories, there are three types of intent errors:
\begin{compactitem}
\item \emph{Random error (I0)}: the random noisy intent from the same category (\emph{within group error}) or other categories (\emph{between group error}).
\item \emph{Within-group error (I1)}: the noisy intent is from the same group of the real intent, for example, the real intent is \textsf{request\_theater}, but the predicted intent from LU module might be \textsf{request\_moviename}.
\item \emph{Between-group error (I2)}: the noisy intent is from the different group, for example, a real intent \textsf{request\_moviename} might be predicted as the intent \textsf{inform\_moviename}.
\end{compactitem}

\paragraph{Slot-level Error:}
At the slot level, there are four error types:
\begin{compactitem}
\item \emph{Random error} (S0): to simulate the noise that is randomly set to the following three types.
\item \emph{Slot deletion} (S1): is to simulate the scenario where the slot is not recognized by the LU component.
\item \emph{Incorrect slot value} (S2): is to simulate the scenario where the slot name is correctly recognized, but the slot value is wrong, e.g., wrong word segmentation.
\item \emph{Incorrect slot} (S3): is to simulate the scenario where both the slot and its value are incorrectly recognized.
\end{compactitem}

\section{End-to-End Reinforcement Learning}
\label{sec:e2e_rl}
To learn the interactive policy of our system, we apply reinforcement learning to the DM training in an end-to-end fashion, where each neural network component can be fine tuned. The policy is represented as a deep Q-network (DQN)~\cite{Mnih15Human}, which takes the state $s_t$ from the state tracker as input, and outputs $Q(s_t, a; \theta)$ for all actions $a$. Two important DQN tricks, target network usage and experience replay are applied, where the experience replay strategy is changed for the dialogue setting.

During training, we use $\epsilon$-greedy exploration and an experience replay buffer with dynamically changing buffer size. At each simulation epoch, we simulate $N$ ($N = 100$) dialogues and add these state transition tuples ($s_t, a_t, r_t, s_{t+1}$) to the experience replay buffer for training. In one simulation epoch, the current DQN will be updated multiple times (depending on the batch size and the current size of experience replay buffer). At the last simulation epoch, the target network will be replaced by the current DQN, the target DQN network is only updated for once in one simulation epoch.

The experience replay strategy is critical for RL training~\cite{schaul2015prioritized}. In our buffer update strategy, we accumulate all experience tuples from the simulation and flush the pool till the current RL agent reaches a success rate threshold (i.e., a threshold which is equal to the performance of a rule-based agent), and then use the experience tuples from the current RL agent to re-fill the buffer. The intuition is that the initial performance of the DQN is not strong enough to generate good experience replay tuples, thus we do not flush the experience replay pool till the current RL agent can reach a certain success rate (for example, the success rate of a rule-based agent). In the rest of the training process, at every simulation epoch, we estimate the success rate of the current DQN agent (by running it multiple dialogues on simulated users). If the current DQN agent is better than the target network, the experience replay buffer will be flushed. 

\begin{figure*}[htb]
\begin{minipage}[b]{.5\linewidth}
\centering
\centerline{\includegraphics[width=\linewidth]{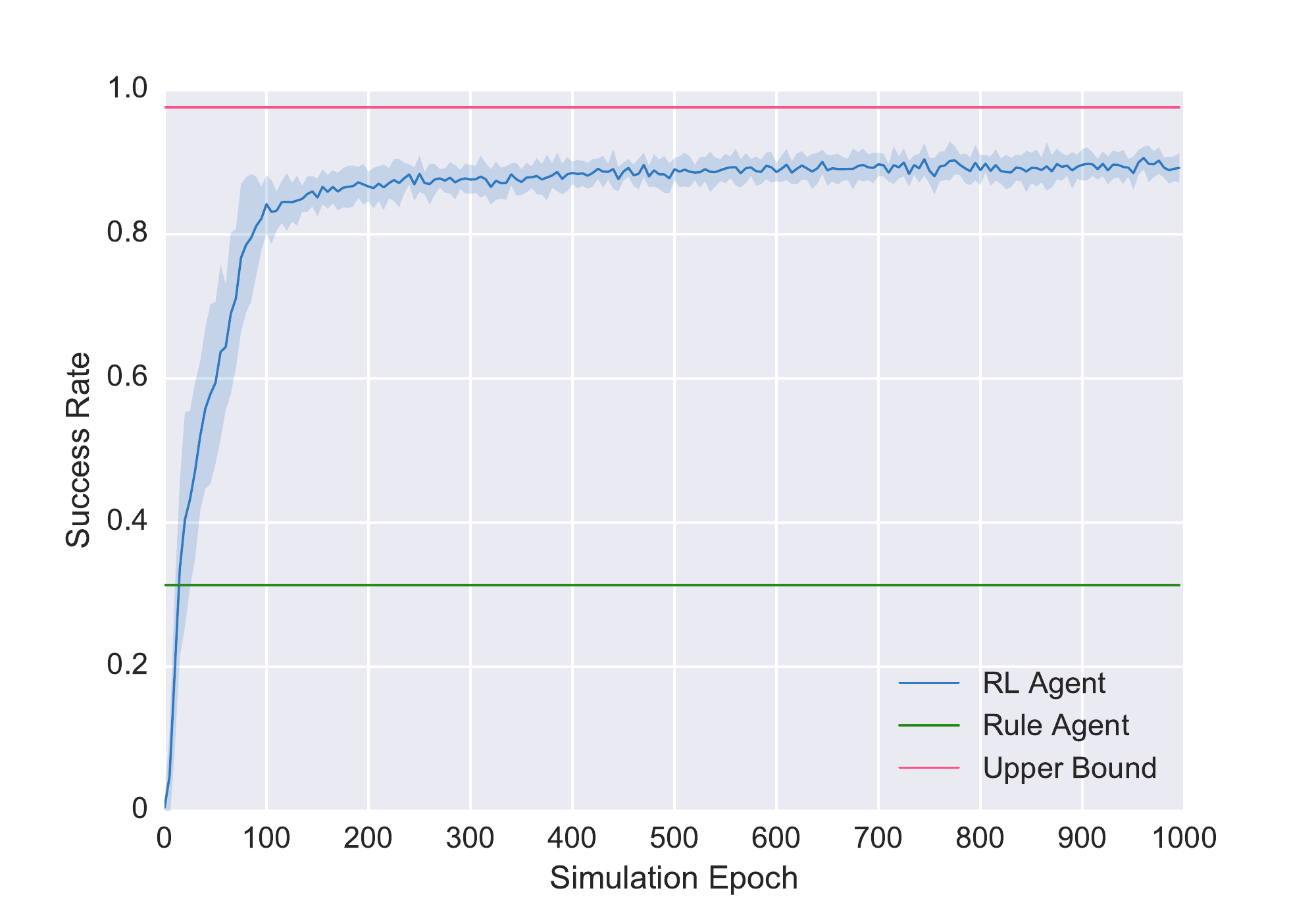}}
\centerline{(a) Frame-level semantics for training}
\end{minipage}
\hfill
\begin{minipage}[b]{0.5\linewidth}
\centering
\centerline{\includegraphics[width=\linewidth]{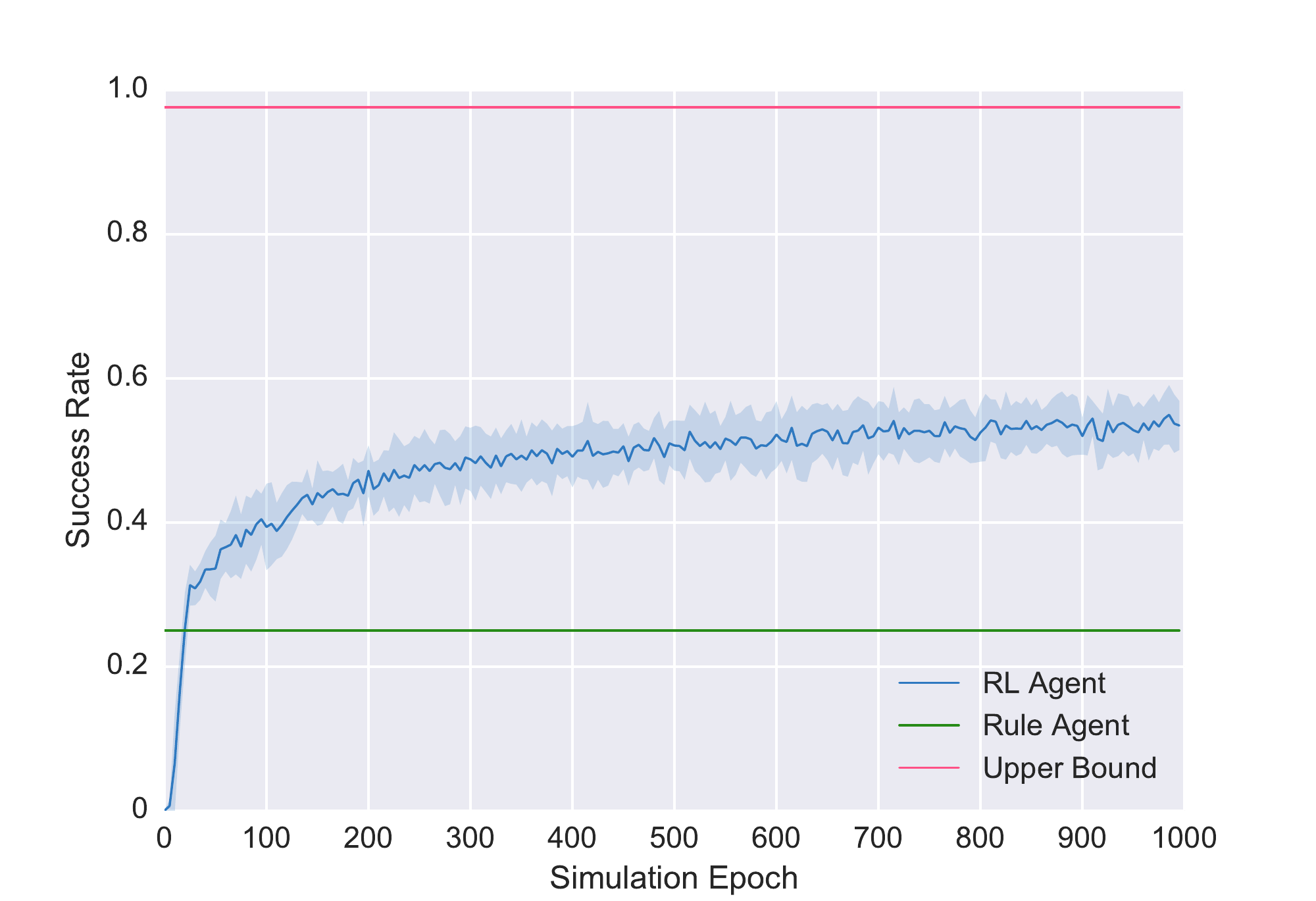}}
\centerline{(b) Natural language for end-to-end training}
\end{minipage}
\vspace{-3mm}
\caption{Learning curves for policy training (average of 10 runs). The blue solid lines show the rule agent performance, where we employ to initialize the experience replay buffer pool; the orange dotted line is the optimal upper bound, which is the percentage of reachable user goals.}
\label{fig:rl_learning_curve}
\end{figure*}

\section{Experiments}
\label{sec:exp}
We consider a task-completion dialogue system for helping users book movie tickets. Over the course of conversation, the dialogue system gathers information about the customer's desires and ultimately books the movie tickets. The environment then assesses a binary outcome (success or failure) at the end of the conversation, based on (1) whether a movie is booked, and (2) whether the movie satisfies the user’s constraints.  

\paragraph{Dataset:}
The raw conversational data were collected via Amazon Mechanical Turk, with annotations provided by domain experts. In total, we have labeled $280$ dialogues, and the average number of turns per dialogue is approximately $11$. The annotated data includes $11$ dialogue acts and $29$ slots, most of the slots are \textit{informable} slots, which users can use to constrain the search, and some are \textit{requestable} slots, of which users can ask values from the agent. For example, \emph{numberofpeople} cannot be a requestable slot, since arguably user knows how many tickets he or she wants to buy. The detailed annotations can be found in Appendix~\ref{app:data}.

\subsection{Simulated User Evaluation}
\label{sec:e2e_train_exp}

Two sets of experiments are conducted in the DM training, where two input formats are used for training the RL agents:
\begin{compactenum}
\item \emph{frame-level semantics}: when training or testing a policy based on semantic frames of user actions, a noise controller described in section~\ref{ssec:user_error}
is used to simulate LU errors and noisy communications between the user and the agent.
\item \emph{natural language}: when training or testing a policy on natural language level, in which LU and NLG may introduce noises. 
In our experiments, the NLG decoder uses $\textit{beam\_size} = 3$ to balance speed and performance.
\end{compactenum}

\begin{table}[t]
\small
  \centering
  \begin{tabular}{cclclc}
    \toprule
    \multicolumn{2}{c}{\multirow{2}{*}{Setting}} & \multicolumn{2}{c}{\textbf{Intent Error}} & \multicolumn{2}{c}{\textbf{Slot Error}} \\
\cline{3-4} \cline{5-6}
& & Type & Rate & Type & Rate \\ \hline
    \multirow{3}{*}{\begin{sideways}Basic\end{sideways}} & {B1} & \multirow{3}{*}{0: random} & 0.00 & \multirow{3}{*}{0: random} & 0.00  \\
    & {B2} &  & 0.10 &  & 0.10 \\
    & {B3} &  & 0.20 &  & 0.20 \\ \hline
    \multirow{6}{*}{\begin{sideways}Intent\end{sideways}} & {I0} & \bf 0: random & 0.10 &  \multirow{6}{*}{0: random} &  \multirow{6}{*}{0.05} \\
    & {I1} & \bf 1: within group & 0.10 &  &  \\
    & {I2} & \bf 2: between group & 0.10 &  &  \\
    & {I3} & 0: random & \bf 0.00 & &  \\
    & {I4} & 0: random & \bf 0.10 &  &  \\
    & {I5} & 0: random & \bf 0.20 &  &  \\ \hline
    \multirow{7}{*}{\begin{sideways}Slot\end{sideways}} & {S0} & \multirow{7}{*}{0: random} & \multirow{7}{*}{0.10} & \bf  0: random & 0.10 \\
    & {S1} &  &  & \bf 1: deletion & 0.10 \\
    & {S2} &  &  & \bf 2: value & 0.10 \\
    & {S3} &  &  & \bf 3: slot & 0.10 \\
    & {S4} &  &  & 0: random & \bf 0.00 \\
    & {S5} &  &  & 0: random & \bf 0.10 \\
    & {S6} &  &  & 0: random & \bf 0.20 \\  
    \bottomrule
  \end{tabular}
\caption{Experimental settings with different intent/slot error types described in section~\ref{ssec:user_error} and different error rates.}
\label{tab:exp_setting}
\end{table}

Figure~\ref{fig:rl_learning_curve}(a) shows a learning curve for the dialogue system performance trained with the frame-level information (user semantic frames and system actions), where the number is the average of 10 runs. Figure \ref{fig:rl_learning_curve}(b) is a learning curve for the system trained at the natural language level. In both settings, the RL agents significantly outperform the rule-based systems, showing the potential of a neural dialogue system that can perform real-world tasks and be improved autonomously through interactions with users. Also, the end-to-end system in Figure~\ref{fig:rl_learning_curve}(b) takes longer for the RL agent to adapt to the noises from LU and NLG, indicating the difficulty of maintaining the system robustness. The consistently increasing trend of our proposed end-to-end system also suggests greater robustness in noisy, real-world scenarios. To further investigate and understand the real impact of the LU component to the robustness of RL agent in the dialogue system, we conduct a series of experiments under different error settings (intent and slot errors from LU) summarized in Table~\ref{tab:exp_setting}, where the learning curves are averaged over $10$ runs.

\begin{figure}[t]
\begin{minipage}[b]{\linewidth}
\centering
\centerline{\includegraphics[width=\linewidth]{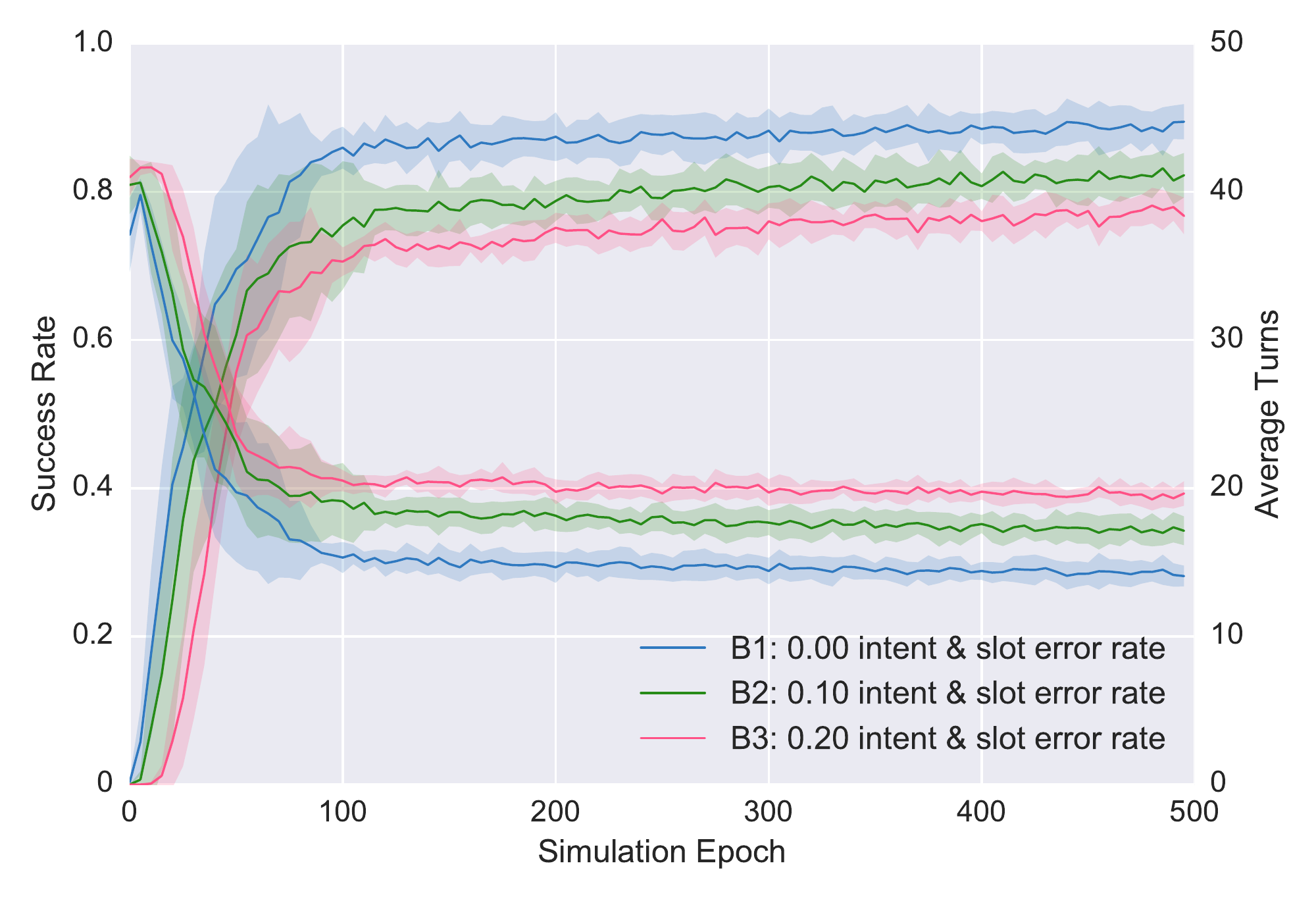}}
\end{minipage}
\caption{Learning curves for different LU error rates.}
\label{fig:basic_learning_curves}
\end{figure}

\begin{figure*}[htb]
\begin{minipage}[b]{.5\linewidth}
\centering
\centerline{\includegraphics[width=\linewidth]{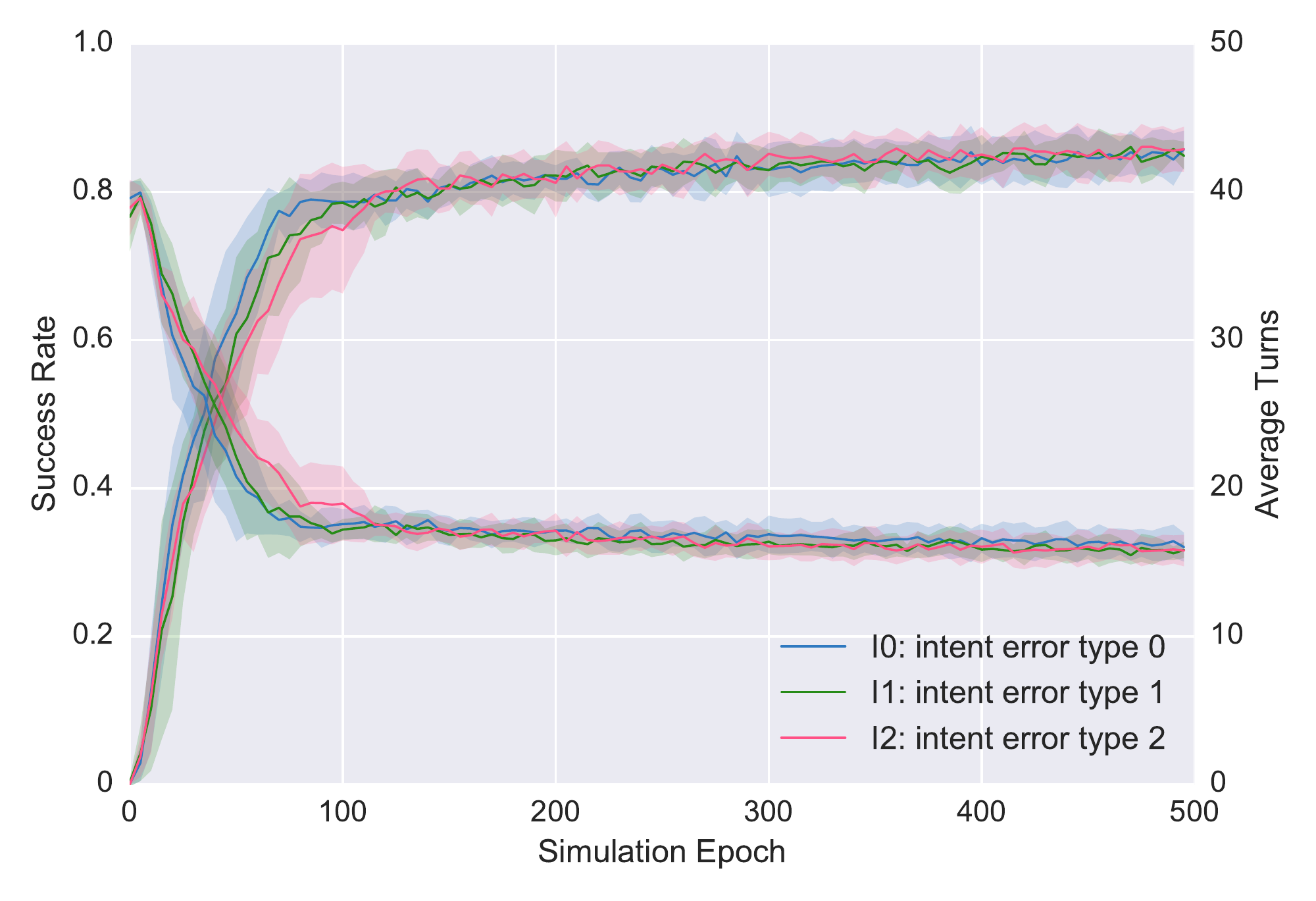}}
\centerline{(a) Intent Error Type Analysis}
\end{minipage}
\hfill
\begin{minipage}[b]{.5\linewidth}
\centering
\centerline{\includegraphics[width=\linewidth]{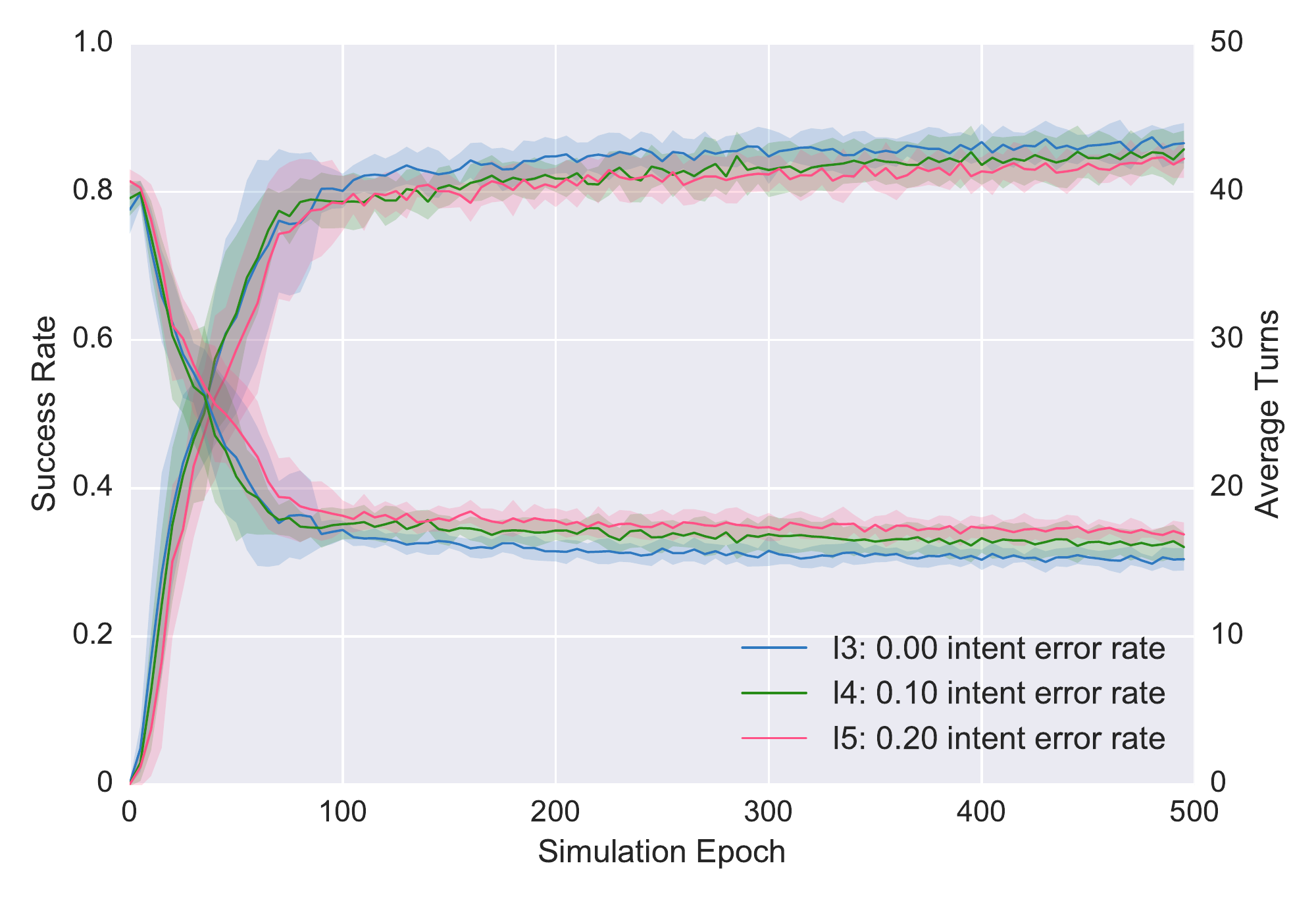}}
\centerline{(b) Intent Error Rate Analysis}
\end{minipage}
\begin{minipage}[b]{.5\linewidth}
\centering
\centerline{\includegraphics[width=\linewidth]{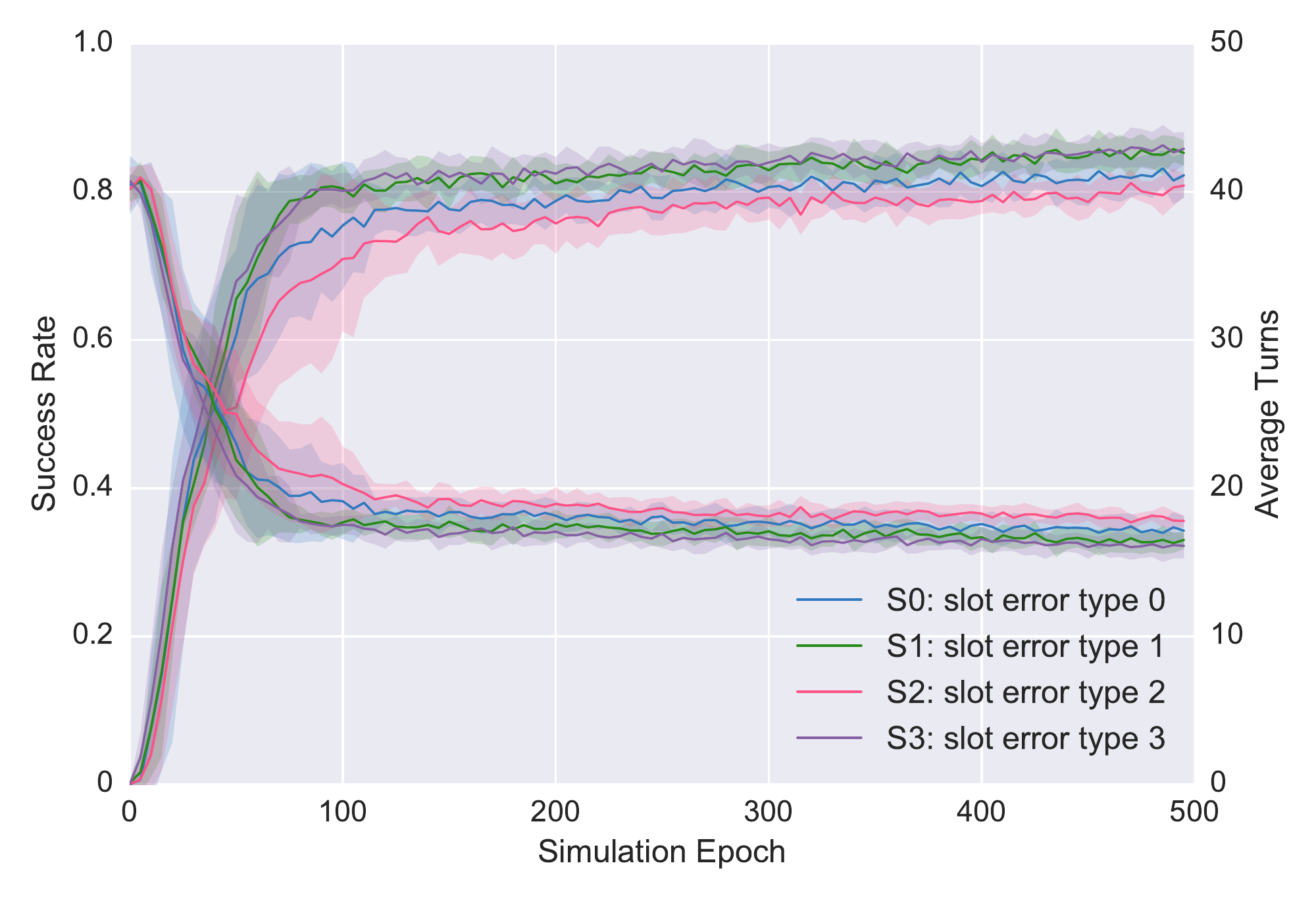}}
\centerline{(c) Slot Error Type Analysis}\medskip
\end{minipage}
\hfill
\begin{minipage}[b]{0.5\linewidth}
\centering
\centerline{\includegraphics[width=\linewidth]{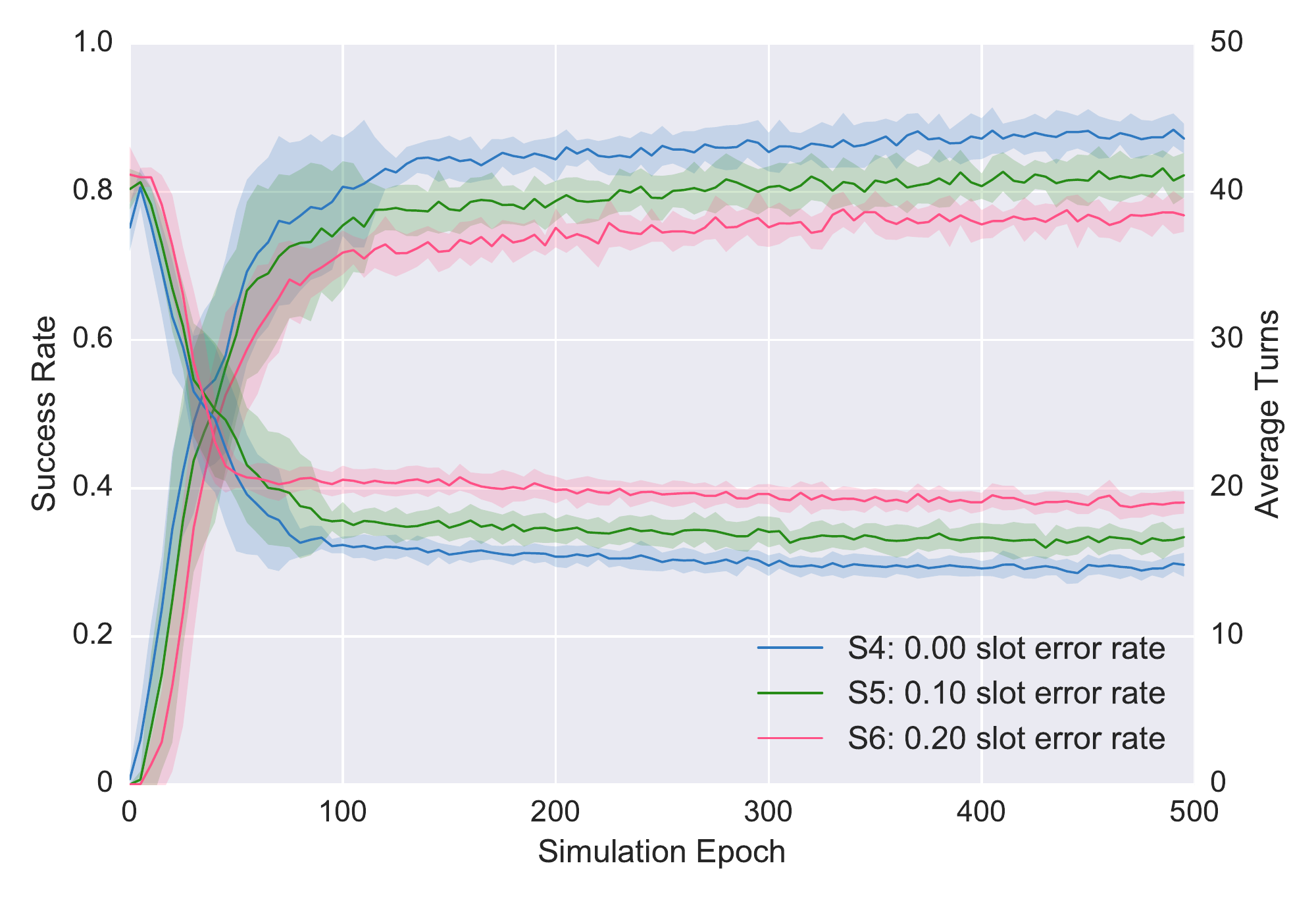}}
\centerline{(d) Slot Error Rate Analysis}
\medskip
\end{minipage}
\vspace{-3mm}
\caption{Learning curves of the different intent and slot errors in terms of success rate (left axis) and average turns (right axis).}
\label{fig:learning_curve}
\end{figure*}

\subsection{Basic Error Analysis}
\label{sec:basic_exp}
The group of basic experiments (from B1 to B3) are in the settings that combine the noise from both intent and slot: 1) For both intent and slot, the error types are random, and the error rates are in $\{0.00, 0.10, 0.20\}$. The rule-based agent reports $41\%$, $21\%$, and $12\%$ success rates under $0.00$, $0.10$, and $0.20$ error rates respectively. In contrast, the RL-based agent achieves $90\%$, $79\%$, and $76\%$ success rate under the same error rates, respectively. We compare the performance between two types of agents and find that the RL-based agent has greater robustness and is less sensitive to noisy inputs. Therefore, the following experiments are performed using a RL dialogue agent due to robustness consideration. From Fig.~\ref{fig:basic_learning_curves}, the dialogue agents degrade remarkably when the error rate increases (leading to lower success rates and higher average turns).

\subsection{Intent Error Analysis}
\label{sec:intent_exp}
To further understand the impact of intent-level noises to dialogue systems, two experimental groups are performed: the first group (I0--I2) focuses on the difference among all intent error types; the second group (I3--I5) focuses on the impact of intent error rates. Other factors are identical for the two groups, with the random slot error type and a $5\%$ slot error rate.

\subsubsection{Intent Error Type}
\label{sec:intent_type_exp}
Experiments with the settings of I0--I2 are under the same slot errors and same intent error rate ($10\%$), but with different intent error types: I1 includes the noisy intents from the same categories, I2 includes the noisy intents from different categories, and I0 includes both via random selection. Fig.~\ref{fig:learning_curve}(a) shows the learning curves for all intent error types, where the difference among three curves is insignificant, indicating that the incorrect intents have similar impact no matter what categories they belong to.

\subsubsection{Intent Error Rate}
\label{sec:intent_err_exp}
Experiments with the settings I3--I5 investigate the difference among different intent error rates. When the intent error rate increases, the dialogue agent performs slightly worse, but the difference is subtle. It suggests that the RL-based agent has better robustness to noisy intents. As shown in Fig.~\ref{fig:learning_curve}(a,b), all RL agents can converge to a similar success rate in both intent error type and intent error rate settings.

\subsection{Slot Error Analysis}
\label{sec:slot_exp}
We further conducted two groups of experiments to investigate the impact of slot-level noises
where other factors are fixed --- with the random intent error type and a $10\%$ intent error rate.

\subsubsection{Slot Error Type}
\label{sec:slot_type_exp}
Experiments (S0 -- S3) investigate the impact of different slot error types.
Corresponding learning curves are given in Fig.~\ref{fig:learning_curve}(c).
Among single error types (S1--S3), \emph{incorrect slot value} (S2) performs worst, which means that the slot name is recognized correctly, but a wrong value is extracted with the slot (such as wrong word segmentation); in this case, the agent receives a wrong value for the slot, and eventually books a wrong ticket or fails to book it. The probable reason is that the dialogue agent has difficulty identifying the mistakes, 
and using the incorrect slot values for the following dialogue actions could significantly degrade the performance. Between \emph{slot deletion} (S1) and \emph{incorrect slot} (S3), the difference is limited, indicating that the RL agent has similar capability of handling these two kinds of slot-level noises.

\subsubsection{Slot Error Rate}
\label{sec:slot_err_exp}
Experiments with the settings from S4 to S6 focus on different slot error rates ($0\%$, $10\%$, and $20\%$) and report the results in Fig.~\ref{fig:learning_curve}(d).
It is clear from Fig.~\ref{fig:learning_curve}(d) that the dialogue agent performs worse as the slot error rate increases (the curve of the success rate drops and the curve of average turns rises). Comparing with Fig.~\ref{fig:learning_curve}(b), the dialogue system performance is more sensitive to the slot error rate than the intent error rate.

\subsection{Human Evaluation}
\label{sec:human_eval}
\begin{figure}[t]
\begin{minipage}[b]{0.495\linewidth}
\centering
\centerline{\includegraphics[width=\linewidth]{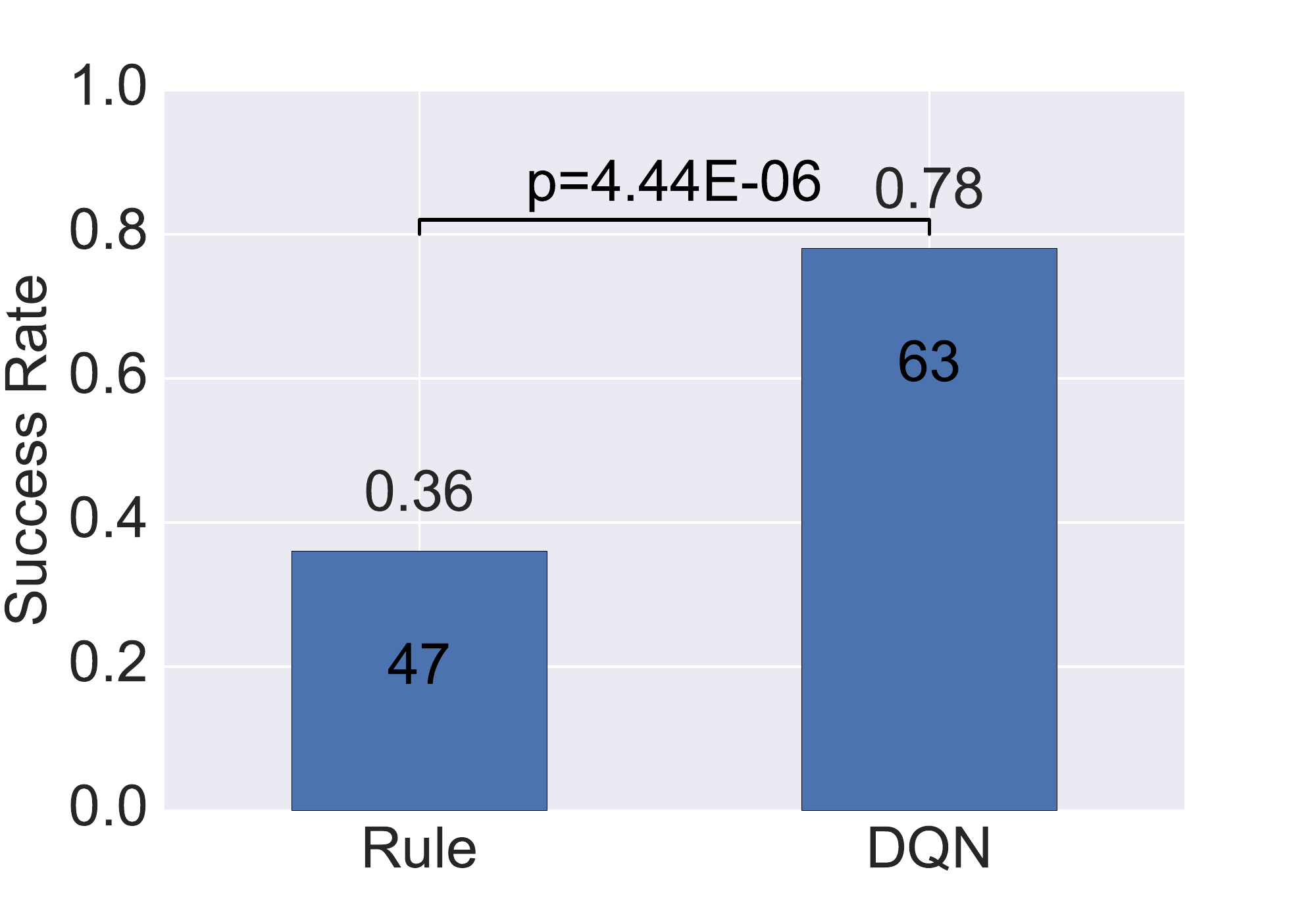}}
\centerline{\small (a) Success Rate}
\end{minipage}
\begin{minipage}[b]{0.495\linewidth}
\centering
\centerline{\includegraphics[width=\linewidth]{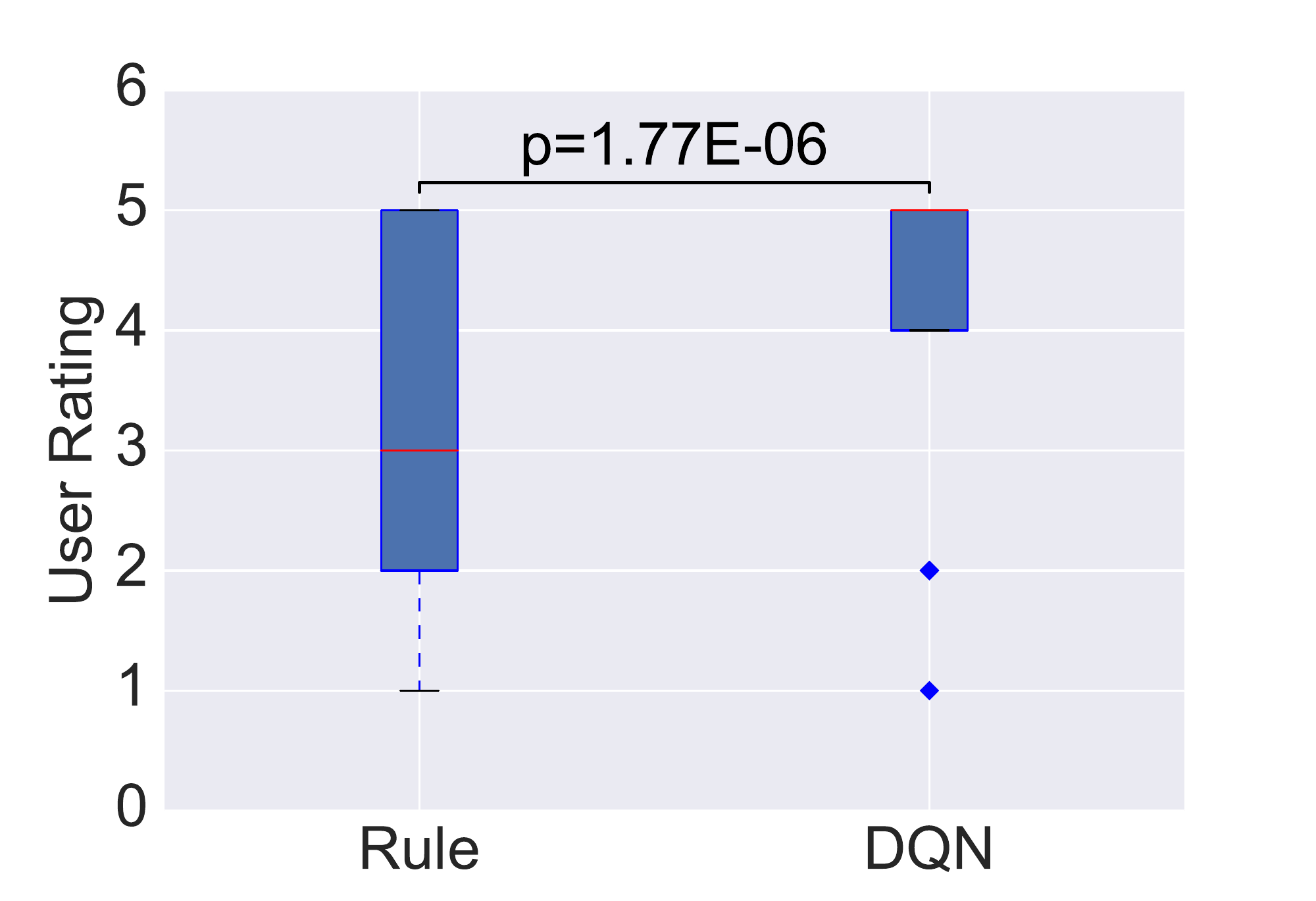}}
\centerline{\small (b) User Rating Distribution}
\end{minipage}
\caption{Performance of the rule-based agent versus DQN agent tested with real users: (a) success rate, number of tested dialogues and p-values are indicated on each bar; (b) Distribution of user ratings for two agents (difference in mean is significant with $p < 0.01$).} 
\label{fig:human_eval_results}
\end{figure}

We further evaluated the rule-based and DQN agents against real human users recruited from the authors’ affiliation, where the DQN agent was trained on the simulated user in the frame-level with $5\%$ random slot errors.
In each dialogue session, one of the agents was randomly picked to converse with a user, and the user was presented with a predefined user goal sampled from our corpus, and was instructed to converse with the agent to complete the presented task.
At the end of each session, the user
was asked to give a rating on a scale from 1 (worst) to 5 (best) based on both \emph{naturalness} and \emph{coherence} of the dialogue.
We collected a total of 110 dialogue sessions from 8 human users.
Figure~\ref{fig:human_eval_results}(a) presents the performance of these agents against real users in terms of success rate. Figure~\ref{fig:human_eval_results}(b) shows the subjective evaluation in terms of user rating.
For all the cases, the RL agent significantly outperforms the rule-based agent for both objective (success rate) and subjective evaluation (user rating). 

\section{Discussion and Future Work}
This paper presents an end-to-end learning framework for task-completion neural dialogue systems. Our experiments, both on simulated and real users, show that reinforcement learning systems outperform rule-based agents and have better robustness to allow natural interactions with users in real-world task-completion scenarios. Furthermore, we conduct a series of extensive experiments to understand the impact of natural language understanding errors on the performance of a reinforcement learning based, task-completion neural dialogue system. Our empirical results suggest several interesting findings: 1) slot-level errors have a greater impact than intent-level errors; A possible explanation is related to our dialogue action representation, \emph{intent(slot-value pairs)}. If an intent is predicted wrong, for example, \emph{inform} was predicted incorrectly as \emph{request\_ticket}, the dialogue agent can handle this unreliable situation and decide to make confirmation in order to keep the correct information for the following conversation. In contrast, if a slot \emph{moviename} is predicted wrong, or a slot value is not identified correctly, this dialogue turn might directly pass the wrong information to the agent, which might lead the agent to book a wrong ticket. Another reason is that the dialogue agent can still maintain a correct intent based on slot information even though the predicted intent is wrong. In order to verify the hypotheses, further experiments are needed, which we leave as future work. 2) different slot error types have different impacts on the RL agents. 3) RL agents are more robust to certain types of slot-level errors --- the agents can learn to double-check or confirm with users, at the cost of slightly longer conversations.

Finally, it should be noted that the experiments in this paper focus on task-completion dialogues. Another type of dialogues known as chit-chats has different optimization goals~\cite{li2016deep}. It would be interesting to extend our analysis from this paper to chit-chat dialogues to gain useful insights for impacts of LU errors.

\section*{Acknowledgments}

We would like to thank Dilek Hakkani-T\"{u}r and reviewers for their insightful comments on the paper.
Yun-Nung Chen is supported by the Ministry of Science and Technology of Taiwan and MediaTek Inc..

\bibliography{ijcnlp2017}
\bibliographystyle{ijcnlp2017}

\newpage
\appendix

\section{Dataset Annotation}
\label{app:data}

Table \ref{tab:data} lists all annotated dialogue acts and slots in detail.

\begin{table}[h]
\small
\begin{tabular}{|c|l|l|}
\hline
\multicolumn{2}{|c|}{Annotations} \\
\hline\hline
& \textsf{request}, \textsf{inform}, \textsf{deny}, \textsf{confirm\_question},\\ 
Intent & \textsf{confirm\_answer}, \textsf{greeting}, \textsf{closing}, \textsf{not\_sure},\\
 & \textsf{multiple\_choice}, \textsf{thanks}, \textsf{welcome} \\
\hline
\multirow{8}{*}{Slot} & \textsf{actor}, \textsf{actress}, \textsf{city}, \textsf{closing}, \textsf{critic\_rating}, \\ 
& \textsf{date}, \textsf{description}, \textsf{distanceconstraints},\\
& \textsf{greeting}, \textsf{implicit\_value}, \textsf{movie\_series},\\
& \textsf{moviename},  \textsf{mpaa\_rating}, \textsf{numberofpeople}, \\
& \textsf{numberofkids}, \textsf{taskcomplete}, \textsf{other}, \textsf{price},\\
& \textsf{seating}, \textsf{starttime}, \textsf{state}, \textsf{theater}, \\ 
& \textsf{theater\_chain}, \textsf{video\_format}, \textsf{zip}, \textsf{result}, \\
& \textsf{ticket}, \textsf{mc\_list} \\
\hline
\end{tabular}
\centering
\caption{The data annotation.}
\label{tab:data}
\end{table}

\section{Sample Dialogues}
Table \ref{tab:tcp_sample} shows one success and one failure dialogue examples generated by the rule-based agent and RL agent interacting with user simulator in the movie-booking domain. To be informative, we also explicitly show the user goal at the head of the dialogue, the purpose is to help the user to accomplish this goal and book the right movie tickets, but the agent knows nothing about the user goal. 

\begin{table*}[htbp!]
\small
\begin{tabular}{l|l}
\hline
\begin{tabular}[c]{@{}l@{}}
New episode, user goal:\\
\{\\
\-\hspace{3mm}  ``request\_slots": \{	\\
\-\hspace{6mm}    ``ticket": ``UNK"	\\
\-\hspace{3mm}  \},	\\
\-\hspace{3mm}  ``inform\_slots": \{	\\
\-\hspace{6mm}    ``city": ``seattle",	\\
\-\hspace{6mm}    ``numberofpeople": ``2",	\\
\-\hspace{6mm}    ``theater": ``regal meridian 16",	\\
\-\hspace{6mm}    ``starttime": ``9:25 pm",	\\
\-\hspace{6mm}    ``date": ``tomorrow",	\\
\-\hspace{6mm}    ``moviename": ``zoolander 2"	\\
\-\hspace{3mm}  \}\\
\}\\
\end{tabular}  & \begin{tabular}[c]{@{}l@{}}
New episode, user goal:\\
\{\\
\-\hspace{3mm} ``request\_slots": \{	\\
\-\hspace{6mm}    ``ticket": ``UNK"	\\
\-\hspace{6mm}    ``theater": ``UNK"	\\
\-\hspace{6mm}    ``starttime": ``UNK"	\\
\-\hspace{3mm}  \},	\\
\-\hspace{3mm}  ``inform\_slots": \{	\\
\-\hspace{6mm}    ``numberofpeople": ``3",	\\
\-\hspace{6mm}    ``date": ``tomorrow",	\\
\-\hspace{6mm}    ``moviename": ``10 cloverfield lane"	\\
\-\hspace{3mm}  \}\\
\}\\
\-\hspace{3mm}
\end{tabular}                \\
\hline
\textbf{Rule-based agent} & \textbf{Rule-based Agent} \\
\hline

\underline{User}: Can I get some tickets for zoolander 2?  & \underline{User}: What is the start time for 10 cloverfield lane?  \\
\underline{Agent}: What movie are you interested in? & \underline{Agent}: What movie are you interested in?  \\
\underline{User}: I want to watch zoolander 2.  & \underline{User}: I want to watch at 10 cloverfield lane.  \\
\underline{Agent}: What time would you like to see it? & \underline{Agent}: What time would you like to see it? \\
\underline{User}: I want to watch at 9:25 pm. & \begin{tabular}[c]{@{}l@{}} \underline{User}: What time is it playing tomorrow? I need 3 tickets. \end{tabular} \\
\underline{Agent}: What city you would like? & \underline{Agent}: What city you would like?  \\
\underline{User}: I want to watch at seattle. & \underline{User}: I do not care.   \\
\underline{Agent}: What date would you like to watch it? &  \underline{Agent}: What date would you like to watch it? \\
\underline{User}: I want to set it up tomorrow. &  \underline{User}: I want to set it up tomorrow.  \\   
\underline{Agent}: Which theater would you like?   &    \underline{Agent}: Which theater would you like?    \\
\underline{User}: I want to watch at regal meridian 16.	&    \underline{User}: Which theater is available?      \\
\underline{Agent}: How many tickets do you need?  &  \underline{Agent}: How many tickets do you need?   \\
\underline{User}: I want 2 tickets please! &  \underline{User}: I want 3 tickets please! \\ 
\begin{tabular}[c]{@{}l@{}}\underline{Agent}: Great - I was able to purchase 2 tickets for you to \\ see zoolander 2 tomorrow at regal meridian 16 theater in \\ seattle at 9:25 pm. \end{tabular}  &  \begin{tabular}[c]{@{}l@{}}\underline{Agent}: Okay, I have booked 3 tickets for 10 cloverfield \\ lane tomorrow. \end{tabular}     \\
\underline{User}: Thank you. & \underline{User}: Thank you. \\ 
\underline{Agent}: Thank you. & \underline{Agent}: Thank you. \\ 
\hline
\textbf{Success} & \begin{tabular}[c]{@{}l@{}} \textbf{Failure:} Agent does not answer all the questions \\ (starttime, theater) from user.\end{tabular}
\\ \hline

\textbf{RL agent} & \textbf{RL Agent} \\
\hline

\underline{User}: Can I get tickets for zoolander 2 tomorrow?  & \begin{tabular}[c]{@{}l@{}}\underline{User}: Which theater can I book 3 tickets for 10 \\ cloverfield lane?  \end{tabular} \\
\underline{Agent}: Which city would you like?    & \underline{Agent}: What time would you like to see it?  \\
\underline{User}: I want to watch at seattle.    & \begin{tabular}[c]{@{}l@{}} \underline{User}: Which theater and start time are available \\ tomorrow?  \end{tabular}\\
\underline{Agent}: How many tickets do you need?  & \underline{Agent}: 11:45am is available.     \\
\underline{User}: I want 2 tickets please!      & \underline{User}: Which theater is available? \\
\underline{Agent}: 9:25 pm is available.       & \underline{Agent}: regal la live stadium 14 is available.  \\
\underline{User}: I want to watch at regal meridian 16.   & \underline{User}: Could you help me to book the tickets?  \\
\begin{tabular}[c]{@{}l@{}}\underline{Agent}: Great - I was able to purchase 2 tickets for you to \\ see zoolander 2 tomorrow at regal meridian 16 theater in \\ seattle at 9:25 pm. \end{tabular}   &   \begin{tabular}[c]{@{}l@{}} \underline{Agent}: Okay - I was able to book 3 tickets for you to see \\ 10 cloverfield lane at regal la live stadium 14 at 11:45am \\ tomorrow. \end{tabular}      \\
\underline{User}: Thank you.    &      \underline{User}:Thank you.  \\   
\underline{Agent}: Thank you.   &      \underline{Agent}: Thank you.     
\\ \hline
\textbf{Success} & \textbf{Success}
\\ \hline

\end{tabular}
\small
\centering
\caption{Two sample dialogues generated by rule-based and RL agents with user simulator: Left column shows both rule and RL agents succeed; Right column shows that rule-based agent fails, while RL agent succeeds.}
\label{tab:tcp_sample}
\end{table*}

\end{document}